\documentclass[journal]{IEEEtran}
\usepackage[hyphens]{url}
\usepackage{hyperref}
\hypersetup{breaklinks=true}
\hypersetup{draft}
\usepackage{gensymb}

\usepackage{comment}
\usepackage{enumitem}
\usepackage{multirow}
\usepackage{float}

\usepackage[margin = 0.8in]{geometry}
\usepackage{nopageno}
\usepackage{cite}
\usepackage{amsmath,amssymb,amsfonts,mathtools}
\usepackage{algorithmic}
\usepackage{graphicx}
\usepackage{textcomp}
\usepackage[table]{xcolor}
\usepackage{listings}
\usepackage{color}

\usepackage{dblfloatfix}    

\ifCLASSINFOpdf
  \usepackage{graphicx}
  \graphicspath{{./Fig/}}
  \DeclareGraphicsExtensions{.pdf,.jpeg,.png}
\else
\fi
\allowdisplaybreaks 

\usepackage{amsmath}
\newcommand\numberthis{\addtocounter{equation}{1}\tag{\theequation}}
\usepackage[linesnumbered,ruled]{algorithm2e}

\SetCommentSty{mycommfont}

\usepackage{caption}

\usepackage{amsthm}
\newtheorem*{assumption*}{\assumptionnumber}
\providecommand{\assumptionnumber}{}
\makeatletter
\newenvironment{assumption}[2]
{%
	\renewcommand{\assumptionnumber}{Assumption #1}%
	\begin{assumption*}%
		\protected@edef\@currentlabel{#1}%
	}
	{%
	\end{assumption*}
}
\makeatother

\usepackage{amsthm}
\newtheorem*{observation*}{\assumptionnumber}
\providecommand{\assumptionnumber}{}


\usepackage{array}
\usepackage{makecell}
\usepackage{blindtext}

\begin{document}
	\bstctlcite{IEEEexample:BSTcontrol}
	
	\raggedbottom
	
%
\title{mmFall: Fall Detection using 4D MmWave Radar and a Hybrid Variational RNN AutoEncoder}
%
%
%

\author{Feng~Jin,~\textit{Student Member, IEEE},~Arindam~Sengupta,~\textit{Student Member, IEEE},~and~Siyang~Cao, \textit{Member, IEEE}
\thanks{Copyright (c) 2015 IEEE. Personal use of this material is permitted. However, permission to use this material for any other purposes must be obtained from the IEEE by sending a request to pubs-permissions@ieee.org.}
\thanks{Feng Jin, Arindam Sengupta and Siyang Cao are with Department of Electrical and Computer Engineering, \textit{the University of Arizona}, Tucson, AZ, 85719 USA. (e-mail: \{fengjin, sengupta, caos\}@email.arizona.edu)}
\\\vspace*{-1cm}}

%
%

\markboth{Preprint Version.}
{Shell \MakeLowercase{\textit{et al.}}: Bare Demo of IEEEtran.cls for IEEE Journals}
%



\maketitle
\pagestyle{plain}

\begin{abstract}
Elderly fall prevention and detection becomes extremely crucial with the fast aging population globally. In this paper, we propose \textit{mmFall} - a novel fall detection system, which comprises of (i) the emerging millimeter-wave (mmWave) radar sensor to collect the human body's point cloud along with the body centroid, and (ii) a Hybrid Variational RNN AutoEncoder (HVRAE) to compute the anomaly level of the body motion based on the acquired point cloud. A fall is detected when the spike in anomaly level and the drop in centroid height occur simultaneously. The mmWave radar sensor offers privacy-compliance and high sensitivity to motion, over the traditional sensing modalities. However, (i) randomness in radar point cloud and (ii) difficulties in fall collection/labeling in the traditional supervised fall detection approaches are the two major challenges. To overcome the randomness in radar data, the proposed HVRAE uses variational inference, a generative approach rather than a discriminative approach, to infer the posterior probability of the body's latent motion state every frame, followed by a recurrent neural network (RNN) to summarize the temporal features over multiple frames. Moreover, to circumvent the difficulties in fall data collection/labeling, the HVRAE is built upon an autoencoder architecture in a semi-supervised approach, which is only trained on the normal activities of daily living (ADL). In the inference stage, the HVRAE will generate a spike in the anomaly level once an abnormal motion, such as fall, occurs. During the experiment\footnote{All the codes and datasets are shared on GitHub (\url{https://github.com/radar-lab/mmfall}).}, we implemented the HVRAE along with two other baselines, and tested on the dataset collected in an apartment. The receiver operating characteristic (ROC) curve indicates that our proposed model outperforms baselines and achieves 98\% detection out of 50 falls at the expense of just 2 false alarms.
\end{abstract}

\def\abstractname{Note to Practitioners}
\begin{abstract}
Traditional non-wearable fall detection approaches typically make use of a vision-based sensor, such as camera, to monitor and detect fall using a classifier that is trained in a supervised fashion on the collected fall and non-fall data. However, several problems render these methods impractical. Firstly, camera-based monitoring may trigger privacy concerns. Secondly, fall data collection using human subjects is difficult and costly, not to mention the impossible ask of the elderly repeating simulated falls for data collection. In this paper, we propose a new fall detection approach to overcome these problems by (i) using a palm-size mmWave radar sensor to monitor the elderly, that is highly sensitive to motion while protecting privacy; and (ii) using a semi-supervised anomaly detection approach to circumvent the fall data collection. Further hardware engineering and more training data from people with different body figures could make the proposed fall detection solution even more practical.
\end{abstract}

\begin{IEEEkeywords}
Fall detection, millimeter wave radar, variational autoencoder, recurrent autoencoder, semi-supervised learning, anomaly detection.
\end{IEEEkeywords}

%
\IEEEpeerreviewmaketitle

\section{Introduction}
\IEEEPARstart{G}{lobally}, the elderly aged 65 or over make up the fastest-growing age group \cite{ref_aging}. Approximately 28-35\% of the elderly fall every year \cite{ref_fallrate}, making it the second leading unintentional injury death after road traffic injuries \cite{ref_fallfact}. Moreover, elderly falls are cost intensive with the total 2015 direct cost of fall among the elderly, adjusted for inflation, being 31.9 billion USD in the United States alone \cite{ref_fallcost}. Therefore, researchers seek to detect the fall right after it occurs, along with an immediate alert trigger so a timely treatment can be implemented \cite{ref_fallreview1}. Based on the choice of sensor, fall-related research can be broadly divided into wearables, non-wearables and fusion domains \cite{ref_fallreview1}.

\par In this paper, we are focusing on non-wearable fall detection using the emerging millimeter-wave (mmWave) radar sensor \cite{ref_radarshort}. In short, mmWave radar represent moving objects in a scene as a point cloud in which each point contains the 3D position in space and a 1-D Doppler (radial velocity component) information, thereby resulting in a 4D mmWave radar as referred to in the paper title. 

\par MmWave radar sensor can offer several advantages over the other traditional sensing technologies, viz. (i) non-intrusive and convenience over the wearable solutions \cite{ref_wearable1, ref_wearable2, ref_wearable3} that also need frequent battery recharging; (ii) privacy-compliance over camera \cite{ref_fallcamera}; (iii) high-sensitivity to motion and operationally robust to occlusions, when compared to depth sensors \cite{ref_falldepth}, especially in a complex living environment; (iv) more informative than typical ambient sensors \cite{ref_fallacoustic, ref_fallviberation, ref_fallthermal} which suffer interference from the external environment \cite{ref_fallreview2}; and (v) low-cost, compact and high resolution over the traditional radar counterparts \cite{ref_fallradar}. 

\par The World Health Organization (WHO) \cite{ref_fallrate} defines fall as ``\textit{inadvertently coming to rest on the ground, floor or other lower level, excluding intentional change in position to rest in furniture, wall or other objects}." Therefore, we propose the \textit{mmFall}, in which a generative recurrent autoencoder measures the motion inadvertence or anomaly level based on the mmWave radar point cloud of the body, and the drop of centroid height, which is estimated from the point cloud, indicates the motion of coming to rest on a lower level. Moreover, such a semi-supervised approach can circumvent the difficulties of real-world elderly fall data collection. 

\par The rest of this paper is organized as follows. Section \ref{sec_relatedwork} discusses related radar-based fall detection research and semi-supervised learning approaches. Section \ref{sec_preliminary} introduces all the components that constitute our proposed \textit{mmFall} system, including the principles of mmWave radar sensor, variational inference, variational autoencoder, and recurrent autoencoder. Section \ref{sec_proposedsystem} presents the overall system architecture, a novel data oversampling method and a custom loss function for model training. Section \ref{sec_experiment} shows the experimental evaluation of \textit{mmFall}, compares the performance with two baseline architectures, and discusses the limitations of current research and future work. Finally, Section \ref{sec_conclusion} concludes the paper.

\section{Related Work}\label{sec_relatedwork}
\par Traditionally in radar-based fall detection research, researchers mainly focus on extracting the micro-Doppler \cite{ref_microdoppler} features, i.e., Doppler distribution over time, and then train a classifier that can distinguish fall from non-fall data \cite{ref_RadarCNN2, ref_fallradarreview1, ref_NEWMICRODOPPER1, ref_FMCWDopplerRADARFALL, ref_GAITMDRADAR, ref_NEWFALL1}. However, micro-Doppler features have no spatial information (range and angle), presence of multiple people, other motion sources, and similar motions as fall (such as sitting), can lead to inaccuracies. Jokanovi´c \cite{ref_fallradarreview2} fused information from both the micro-Doppler and range domains to reduce the false alarm rate by training a logistic regression classifier. Similar research can be found in \cite{ref_RANGEDOPPLER1, ref_RANGEDOPPLER2, ref_RANGEDOPPLER3}. Tian \textit{et al.} \cite{ref_MITFALL} used a 3D convolutional neural network (CNN) to learn different activities by exploiting the range-angle heatmap in both azimuth and elevation over time. However, there are two problems in incorporating the spatial information, viz. i) achieving high angular resolution using low-frequency radars requires a bulky antenna (see Fig. 1 of a 4 GHz radar in \cite{ref_RadarCNN2}); ii) and a low signal bandwidth also limits the range resolution. 

\par The use of high-bandwidth mmWave radar looks very promising to overcome these limitations than its traditional counterparts, and is an emerging trend. In our previous research \cite{ref_PatientMonitoring, ref_GROUPWORK1, ref_GROUPWORK2} we adopted such a palm-size mmWave radar sensor that first segregates multiple people based on the spatial information, and then uses a CNN to classify each person's activity, including fall, based on the Doppler pattern separately, and even reconstruct their skeletal pose. Sun \textit{et al.} \cite{ref_MMWAVERADARFALL} also used a mmWave radar and long short-term memory (LSTM) to detect fall based on the range-angle heatmap over time. The advantage of using mmWave radars is furthered if we can take advantage of all the information available from it, such as range, azimuth angle, elevation angle, and Doppler.

\par On the other hand, a vast majority of these radar-based fall detection research adopts a supervised approach. Researchers manually label the collected fall and non-fall data, manually or automatically extract features over time, and then train a classifier that can distinguish fall from non-fall data. The challenge with these supervised approaches is that the rare and non-continuous fall event is very difficult to collect, not to mention the impossible ask of the elderly repeating falls for data collection. Furthermore, the manual extraction and labelling of short portions of fall event from the long duration data is very expensive, time-consuming and inefficient. 

\par To overcome these problems, we leverage the semi-supervised anomaly detection (SSAD) approach. Anomaly detection refers to the problem of finding patterns in data that do not conform to expected behavior \cite{ref_anomalysurvey}. In our case, we can use SSAD to train a model only on the normal activities of daily living (ADL), such as walking/sitting/crouching, etc., such that the model will ‘recognize’ the normal ADL, while a fall event will ‘surprise’ the model as an anomaly.

\par The commonly used SSAD methods include one-class support vector machine (SVM), autoencoders, etc. \cite{ref_SEMIANOMALYDETECTION1, ref_SEMIANOMALYDETECTION2}. SSAD has been applied to detect fall using other sensor modalities \cite{ref_OTHERFALLDETECTIONSEMI1, ref_OTHERSSADFALL1, ref_OTHERSSADFALL2, ref_OTHERSSADFALL3}. However, we found little research on SSAD in radar-based fall detection systems. Diraco \textit{et al.} \cite{ref_RADARSEMI3} introduced one-class SVM and the K-means based approach using micro-Doppler features obtained from a 4.3 GHz radar. As normal ADL data is generally easy to collect and imminently develop, we prefer to use autoencoders \cite{ref_AEREVIEW}, which, unlike other methods, can be incorporated into a neural network to learn large-scale datasets.

\par Particularly, we propose a Hybrid Variational RNN AutoEncoder (HVRAE) that adopts two autoencoder substructures, viz. i) the variational (inference) autoencoder (VAE) \cite{ref_vae}, a generative model rather than a discriminative model, to learn the radar data per frame; ii) a recurrent autoencoder (RAE) to learn temporal features over multiple frames to model fall as a sequence of events. Combining the VAE and RAE has been widely studied in computer vision (CV) and natural language processing (NLP) areas. Fabius \textit{et al.} \cite{ref_ORIGINALVRAE} developed the Variational Recurrent Autoencoder (VRAE) that first uses RAE to summarize the temporal features over multiple frames and then uses VAE to learn the distribution of the summarized features. Chung \textit{et al.} \cite{ref_ORIGINALVRNN} proposed a more profound structure called Variational RNN (VRNN) that applies VAE every frame to learn the distribution but conditioned on previous frame. Our HVRAE model, in which VAE performs on every frame independently, can be viewed as an adaptation of VRNN to simplify the temporal learning. Similar models have been proposed to detect anomaly in other applications as outlined in \cite{ref_LSTMAE2}.

\par In summary, our major contributions include: i) the first method to detect fall based on the 4D radar point cloud of a human in a semi-supervised approach; ii) introducing a variational inference into radar point cloud distribution learning.

\section{Preliminaries}\label{sec_preliminary}
In this section, we introduce the background of all the components that constitute the proposed \textit{mmFall} system detailed in the next section.
\subsection{4D mmWave FMCW Radar Sensor}\label{sec_radar}
\par The carrier frequency of mmWave frequency-modulated continuous-wave (FMCW) radar sensor, or mmWave radar sensor for short, ranges from 57 GHz to 85 GHz according to various applications. For example, 76-81 GHz is primarily used for automotive applications such as objects' dynamics measurement \cite{ref_76_81GHz}, and 57-64 GHz can be used for short-range interactive motion sensing such as in Google's Soli project \cite{ref_57_64GHz}. Coming along with the high carrier frequency, a high bandwidth up to 4 GHz is available, and the physical size of hardware components, including antennas, shrinks. This eventually makes the mmWave radar sensor more compact and higher resolution than the traditional low-frequency band radars. 
\par There are no significant differences in signal modulation and processing of mmWave radar sensor than that of conventional FMCW radars described in \cite{ref_radasignalprocessing}. Generally, the mmWave radar sensor transmits multiple linear FMCW signals over multiple antenna channels in both azimuth and elevation. After the stretch processing and digitalization, a raw multidimensional radar data cube is obtained. Followed by a series of fast Fourier transform (FFT), the parameters of each reflection point in a scene, i.e., range $r$, azimuth angle $\theta_{AZ}$, elevation angle $\theta_{EL}$, and Doppler $DP$, are estimated. In addition, during this process the constant false alarm rate (CFAR) is incorporated to detect the points with signal-to-noise ratio (SNR) greater than an adaptive threshold, and the moving target indication (MTI) is applied to distinguish the moving points from the static background. Eventually, a set of moving points, also called radar point cloud, is obtained.
\begin{figure}[H]
	\centering
	\includegraphics[scale=0.15]{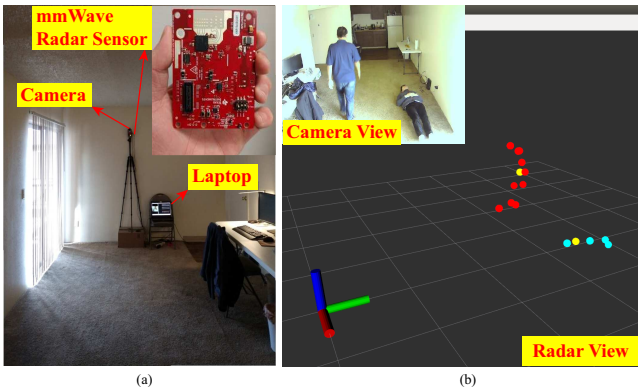}
	\vspace*{-0.1cm}
	\caption{MmWave radar sensor and radar point cloud. (a) The mmWave radar sensor is set up in an apartment, the camera provides a view for reference, and the laptop is used for data acquisition. The same setup is also used in the experiment in Section \ref{sec_experiment}. (b) Radar point cloud in a two-person scenario (lying down on the floor and walking). For the points, different color indicates different person while the yellow point indicates the centroid. For the coordinates, red is the cross-radar direction, green is the forward direction, and blue is the height direction. The original radar measurement of each point is a vector of $(r, \theta_{\text{AZ}}, \theta_{\text{EL}}, {DP})$, along with the estimated centroid of $(x_c, y_c, z_c)$.}
	\vspace*{-0.2cm}
	\label{fig_radarsummary}
\end{figure}
\par If multiple moving targets are present in a scene, the obtained point cloud is a collection of such points from all targets. Thus, a clustering method, such as the Density-Based Spatial Clustering of Applications with Noise (DBSCAN), has to be applied to segregate multiple targets. Meanwhile, the target's centroid can be estimated from the point subset associated with it. Followed by a tracking algorithm, such as Kalman filtering, the trajectory of each target will be recorded with an association of a unique target ID. Particularly, a joint clustering/tracking algorithm called Group Tracking \cite{ref_blackmanMTT} can be used as well. Fig. \ref{fig_radarsummary} shows an example of the mmWave radar sensor and the radar point cloud. With the help of target ID, the motion history of each people can be gathered separately, such that we are able to analyze each person's motion individually. For simplicity but without loss of generality, we will only discuss the single-person scenario thereafter.

\subsection{Radar Point Cloud Distribution for Human Body Motion}\label{sec_radarpointcloudmodel}
\par From Fig. \ref{fig_radarsummary} (b), a straightforward fall detection approach could be to analyze the height of the body centroid. For instance, a fall can be detected when there is a sudden drop in the body centroid. However, this approach may easily cause a false alarm when the person is crouching or sitting.
\par
\begin{figure}[H]
	\centering
	\scalebox{0.5}{
		\input{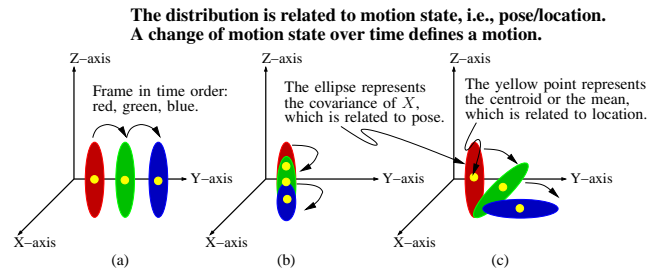}}
	\vspace*{-0.1cm}
	\caption{A depiction of motion pattern. Compare this figure with Fig. \ref{fig_radarsummary} (b). (a) Walking; (b) Crouching; (c) Fall.}
	\vspace*{-0.2cm}
	\label{fig_depiction_motionpattern}
\end{figure}

 Considering the randomness in radar measurement, we now start to view the radar point cloud of the human body as a probabilistic distribution. From the observation in Fig. \ref{fig_radarsummary} (b), the distribution of the point cloud of the lying-down person is different than that of the walking person. Specifically, the covariance of the distribution is related to the human pose, and the mean is related to the human body centroid's location. Therefore, a distribution point of view has a physical significance, in a way that it represents the human pose and location of a person. Moreover, a motion, such as walking/fall, is a change of human pose/location over time, and we therefore call the pose/location as "motion state" for short. A depiction of motions is shown in Fig. \ref{fig_depiction_motionpattern}.
From the discussion above, now we make an assumption
\begin{assumption}{1}{}\label{as_motionpattern}
Let $\mathbf{X}$ denote the radar point cloud of the human body. Let $\mathbf{z}$ denote the body's motion state representing the pose and location. The assumption is, given a $\mathbf{z}$, the distribution of $\mathbf{X}$, i.e., the likelihood $p(\mathbf{X|z})$, follows a particular multivariate Gaussian distribution. And a change of $\mathbf{z}$ over multiple frames defines a motion, such as walking or fall, etc., we therefore need to infer $p(\mathbf{z|X})$ at every frame and learn the change of $\mathbf{z}$ over multiple frames.
\end{assumption}

\par Although Assumption \ref{as_motionpattern} might not hold true as we never know the true physical generation process of radar data from a human body, we at least believe that this assumption is enough for our purpose, i.e., distinguish different human motion. Therefore, we propose to intuitively detect fall through `learning' the uniqueness of such motion patterns.
\par The overview of following subsections is, we propose to (i) learn the distribution at each frame through variational inference, (ii) learn the distribution change over multiple frames through recurrent neural network (RNN), (iii) and discuss, overall, in the framework of autoencoder for semi-supervised learning approach.

\subsection{Variational Inference}
\par More formally, at each frame we obtain a $N$-point radar point cloud $\mathbf{X}{=}\{\mathbf{x}_n\}_{n=1}^N$. The original radar measurement of each point ${\mathbf{x}_n}$ is a four-dimensional vector of $(r, \theta_{\text{AZ}}, \theta_{\text{EL}}, {DP})$. After coordinates transformation to the Cartesian coordinate, ${\mathbf{x}_n}$ goes to be $(x, y, z, {DP})$. We view the points in $\mathbf{X}$ are independently drawn from the likelihood $p(\mathbf{X}|\mathbf{z})$, given a latent motion state $\mathbf{z}$ which is a $D$-dimensional continuous vector. According to Assumption \ref{as_motionpattern}, $p(\mathbf{X}|\mathbf{z})$ follows multivariate Gaussian distribution. The Bayes' theorem shows
\begin{equation}
\underbrace{p(\mathbf{z}|\mathbf{X})}_\text{posterior}=\frac{\overbrace{p(\mathbf{X}|\mathbf{z})}^\text{likelihood}\overbrace{p(\mathbf{z})}^\text{prior}}{\underbrace{p(\mathbf{X})}_\text{evidence}}=\frac{p(\mathbf{X}|\mathbf{z})p(\mathbf{z})}{\int p(\mathbf{X}|\mathbf{z})p(\mathbf{z})d\mathbf{z}}.
\label{posterior}
\end{equation} 
\par We expect to infer the motion state $\mathbf{z}$ based on the observation $\mathbf{X}$. This is equivalent to infer the posterior $p(\mathbf{z}|\mathbf{X})$ of $\mathbf{z}$. Due to the difficulties in solving $p(\mathbf{z}|\mathbf{X})$ analytically as the evidence $p(\mathbf{X})$ is usually intractable, two major approximation approaches, i.e Markov Chain Monte Carlo (MCMC) and variational inference (VI), are mostly used.
\par Generally, the MCMC approach \cite{ref_patternbook} uses a sampling method to draw enough samples from a tractable proposal distribution which is eventually approximate to the target distribution $p(\mathbf{z}|\mathbf{X})$. The most commonly used MCMC algorithm iteratively samples a data $\mathbf{z}^t$ from an arbitrary tractable proposal distribution $q(\mathbf{z}^t|\mathbf{z}^{(t-1)})$ at step $t$, and then accept it with a probability of 
\begin{align*}
&\operatorname*{min}\{1, \frac{p(\mathbf{z}^t|\mathbf{X})*q(\mathbf{z}^t|\mathbf{z}^{(t-1)})}{p(\mathbf{z}^{(t-1)}|\mathbf{X})*q(\mathbf{z}^{(t-1)}|\mathbf{z}^t)}\}\\
&=\operatorname*{min}\{1, \frac{p(\mathbf{X}|\mathbf{z}^t)p(\mathbf{z}^t)*q(\mathbf{z}^t|\mathbf{z}^{(t-1)})}{p(\mathbf{X}|\mathbf{z}^{(t-1)})p(\mathbf{z}^{(t-1)})*q(\mathbf{z}^{(t-1)}|\mathbf{z}^t)}\},
\numberthis
\label{acceptprob}
\end{align*}
where the difficult calculation of $p(\mathbf{X})$ has been circumvented. And it has been proven that this approach constructs a Markov chain whose equilibrium distribution equals to $p(\mathbf{z}|\mathbf{X})$ and is independent to the initial choice of $q(\mathbf{z}^0)$. One of the disadvantages in the MCMC approach is that the chain needs a long and indeterminable burn-in period to approximately reach the equilibrium distribution. This makes the MCMC not suitable for learning on large-scale dataset.
\par On the other hand, the VI approach \cite{ref_vireview} uses a family of tractable probability distribution $Q\{q(\mathbf{z})\}$ to approximate the true $p(\mathbf{z}|\mathbf{X})$ instead of solving it analytically. The VI approach changes the inference problem to an optimization problem as
\begin{equation}
q^*(\mathbf{z})=\operatorname*{arg\,min}_{q(\mathbf{z})\in Q} {\operatorname*{KLD}}\{q(\mathbf{z})||p(\mathbf{z}|\mathbf{X})\},
\label{vibasic}
\end{equation}
where $\operatorname*{KLD}$ is the Kullback–Leibler divergence that measures the distance between two probability distributions. And by definition we have
\begin{align*}
&\operatorname*{KLD}\{q(\mathbf{z})||p(\mathbf{z}|\mathbf{X})\}\\
&\coloneqq \int q(\mathbf{z})\log\frac{q(\mathbf{z})}{p(\mathbf{z}|\mathbf{X})}d\mathbf{z}\\
&=\int q(\mathbf{z})\log q(\mathbf{z})d\mathbf{z} - \int q(\mathbf{z})\log{p(\mathbf{z}}|\mathbf{X})d\mathbf{z}\\
&=\mathbb{E}_{q}[\log{q(\mathbf{z})}] - \mathbb{E}_{q}[\log{p(\mathbf{z}}|\mathbf{X})]\\
&=\mathbb{E}_{q}[\log{q(\mathbf{z})}] - \mathbb{E}_{q}\{\log{[p(\mathbf{X}}|\mathbf{z})p(\mathbf{z})]\} + \mathbb{E}_{q}[\log{p(\mathbf{X})}]\\
&=\underbrace{\mathbb{E}_{q}[\log{q(\mathbf{z})}] - \mathbb{E}_{q}\{\log{[p(\mathbf{X}}|\mathbf{z})p(\mathbf{z})]\}}_{\mathcal{L}(q)} + \log{p(\mathbf{X})},
\numberthis
\label{KLD}
\end{align*}
where $\mathbb{E}_{q}[*]$ is the statistical expectation operator of function $*$ whose variable follows $q(\mathbf{z})$, and $\mathcal{L}$ is called the evidence low bound (ELBO). As the term $\log{p(\mathbf{X})}$ is constant with respect to $\mathbf{z}$, the optimization in Equ. (\ref{vibasic}) is simplified to be 
\begin{align*}
q^*(\mathbf{z})&=\operatorname*{arg\,min}_{q(\mathbf{z})\in Q} {\mathcal{L}(q)}.
\numberthis
\label{equ_viloss}
\end{align*}
\par Here, the difficult computation of $p(\mathbf{X})$ is also circumvented. This optimization approach leads to one of the advantages of VI, that it can be integrated into a neural network framework and optimized through the back-propagation algorithm.
\par It is critical to choose the variational distribution $Q\{q(\mathbf{z})\}$ such that it is not only flexible enough to closely approximate the $p(\mathbf{z}|\mathbf{X})$, but also simple enough for efficient optimization. The most commonly used option is the factorized Gaussian family
\begin{equation}
q(\mathbf{z})=\prod_{d=1}^Dq(\mathbf{z}[d])=\prod_{d=1}^D\mathcal{N}(\mathbf{z}[d]|\boldsymbol{\mu}_q[d], \boldsymbol{\sigma}_q[d]),
\label{equ_meanfield}
\end{equation}
where $(\boldsymbol{\mu}_q, \boldsymbol{\sigma}_q)$ are mean and covariance of the distribution of latent variable $\mathbf{z}$ with a predetermined length of $D$, and the components in $\mathbf{z}$ are mutually independent.
\subsection{Variational Autoencoder}  
\par
\begin{figure}[H]
	\centering
	\scalebox{0.35}{
		\input{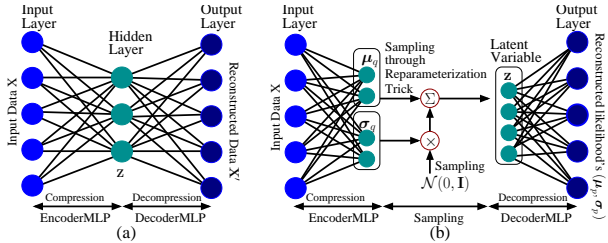}}
	\vspace*{-0.1cm}
	\caption{Autoencoder architecture. (a) Vanilla autoencoder architecture. (b) Variational autoencoder architecture with factorized Gaussian parametrized by $(\boldsymbol{\mu}_q, \boldsymbol{\sigma}_q)$.}
	\vspace*{-0.2cm}
	\label{fig_basicAE}
\end{figure}

 As we briefly state previously, we adopt the semi-supervised anomaly detection approach to train model only on normal ADL such that the model will be surprised by the `unseen' fall data. The common approach is autoencoder, whose basic architecture is shown in Fig. \ref{fig_basicAE} (a). The autoencoder consists of two parts, i.e., encoder and decoder. In most cases, the decoder is simply a mirror of the encoder. The encoder compresses the input data $\mathbf{X}$ to a latent feature vector $\mathbf{z}$ with fewer dimensions, and reversely the decoder reconstructs $\mathbf{X^\prime}$ to be as close to $\mathbf{X}$ as possible, based on the latent feature vector $\mathbf{z}$. Generally, the Multilayer Perceptrons (MLP) are used to model the non-linear mapping function between $\mathbf{X}$ and $\mathbf{z}$, as the MLP is a powerful universal function approximator \cite{ref_HAYKIN2}. Besides a predetermined non-linear activation function, such as sigmoid/tanh, the MLP is characterized by its weights and biases. The training objective is to minimize the loss function between $\mathbf{X}$ and $\mathbf{X^\prime}$ with respect to the weights and biases of encoder MLP and decoder MLP. The loss function could be cross-entropy for a categorical classification problem or mean square error (MSE) for a regression problem.
\par In this way, the autoencoder squeezes the dimensionality to reduce the redundancy of input data. So it learns a compressed yet informative latent feature vector in $\mathbf{X}$. Therefore, the autoencoder will result in a close reconstruction $\mathbf{X^\prime}$ from the input data similar to $\mathbf{X}$, with a low reconstruction loss. However, whenever an `unseen' data passes through, the autoencoder will erroneously squeeze it and be unable to reconstruct it well. This will lead to a loss spike from which an anomaly can be detected.
\par Similarly, in VAE \cite{ref_vae, ref_vae2} in Fig. \ref{fig_basicAE} (b), the encoder learns $q(\mathbf{z})$ that aims to approximate $p(\mathbf{z}|\mathbf{X})$ from the input data $\mathbf{X}$ using VI approach, and the decoder reconstructs the $p(\mathbf{X}|\mathbf{z})$ based on $\mathbf{z}$ sampled from the learned $q(\mathbf{z})$. The VAE training objective is as in Equ. \ref{equ_viloss}. From Equ. \ref{KLD}, the loss function is
\begin{align*}
\mathcal{L}_{\text{VAE}}=\mathcal{L}(q)&=\mathbb{E}_{q}[\log{q(\mathbf{z})}] - \mathbb{E}_{q}[\log p(\mathbf{z})] - \mathbb{E}_{q}[\log{p(\mathbf{X}}|\mathbf{z})]\\
&=\operatorname{KLD}\{q(\mathbf{z})||p(\mathbf{z})\} - \mathbb{E}_{q}[\log{p(\mathbf{X}}|\mathbf{z})].
\numberthis
\label{equ_vaeloss}
\end{align*} 
\par For the variational distribution $q(\mathbf{z})$, the factorized Gaussian in Equ. \ref{equ_meanfield} is used, and for the prior $p(\mathbf{z})$, a common choice of Gaussian $\mathcal{N}(\mathbf{z}|\mathbf{0}, \mathbf{I})$ is used as we do not have a strong assumption on it. Therefore, the first term in $\mathcal{L}_{\text{VAE}}$ in Equ. (\ref{equ_vaeloss}) is reduced to
\begin{align*}
&\operatorname{KLD}\{q(\mathbf{z})||p(\mathbf{z})\}\\
=&-\frac{1}{2}\sum_{d=1}^{D} \{1 + \log\boldsymbol{\sigma}_q[d]^2-\boldsymbol{\mu}_q[d]^2-\boldsymbol{\sigma}_q[d]^2\},
\numberthis
\label{equ_vaeloss_KLD}
\end{align*}
where $(\boldsymbol{\mu}_q, \boldsymbol{\sigma}_q)$ is the mean and variance of the factorized Gaussian $q(\mathbf{z})$ with $D$-dimensional latent vector $\mathbf{z}$. See Appendix \ref{equ8_proof} for detailed derivation.
\par For the second term in $\mathcal{L}_{\text{VAE}}$ in Equ. (\ref{equ_vaeloss}), it is reduced to
\begin{align*}
&\mathbb{E}_{q}[\log p(\mathbf{X}|\mathbf{z})] \\
=& \int q(\mathbf{z}) \log p(\mathbf{X}|\mathbf{z})d\mathbf{z}\\
&(Using \: the \: single{-}data \: Monte \: Carlo \:estimation, \\
&the \: single{-}data \: \mathbf{z} \: is \: sampled \: from \: q(\mathbf{z}), then)\\
\approx& \log p(\mathbf{X}|\mathbf{z})\\
&(Using \: the \: Assumption \: \ref{as_motionpattern} \: that \: likelihood \: follows \: Gaussian)\\
=&\log \mathcal{N}(\mathbf{X}|\boldsymbol{\mu}_p, \boldsymbol{\sigma}_p)\\
=&\log \prod_{n=1}^{N}\mathcal{N}(\mathbf{x}_n|\boldsymbol{\mu}_p, \boldsymbol{\sigma}_p)\\
=&\sum_{n=1}^{N}\sum_{k=1}^{K}\log \mathcal{N}(\mathbf{x}_n[k]|\boldsymbol{\mu}_p[k], \boldsymbol{\sigma}_p[k])\\
&(Ignoring \: the \: constant \: \log \sqrt{2\pi} \: in \: optimization \: problem)\\
\approx&-\frac{1}{2}\sum_{n=1}^{N}\sum_{k=1}^{K}\{\frac{(\mathbf{x}_n[k]-\boldsymbol{\mu}_p[k])^2}{\boldsymbol{\sigma}_p[k]^2} + \log\boldsymbol{\sigma}_p[k]^2\},
\numberthis
\label{equ_vaeloss_logpxz}
\end{align*}
where $\mathbf{X}{=}\{\mathbf{x}_n\}^N_{n=1}$ is the input point cloud, each point $\mathbf{x}_n$ is a $K$-dimensional vector, and $(\boldsymbol{\mu}_p, \boldsymbol{\sigma}_p)$ is the mean and variance of the likelihood $p(\mathbf{X}|\mathbf{z})$.
\par In the third line in Equ. (\ref{equ_vaeloss_logpxz}), a single sample of $\mathbf{z}$ is needed. Instead of drawing from $q(\mathbf{z})$ directly, the reparameterization trick \cite{ref_vae, ref_repara} is used as
\begin{align*}
\mathbf{z}=\boldsymbol{\mu}_p+\boldsymbol{\sigma}_p\odot \boldsymbol{\epsilon},
\numberthis
\label{equ_reparatrick}
\end{align*}
where $\boldsymbol{\epsilon} \sim \mathcal{N}(\mathbf{0},\mathbf{I})$, $\odot$ is element-wise product. The trick is first to draw a sample $\boldsymbol{\epsilon}$ from $\mathcal{N}(\mathbf{0},\mathbf{I})$, and then compute $\mathbf{z}$.
\par By viewing Equ. (\ref{equ_vaeloss_KLD}), the VAE encoder becomes clear as
\begin{equation}
(\boldsymbol{\mu}_q, \log\boldsymbol{\sigma}_q^2)=\operatorname{EncoderMLP}_{\boldsymbol{\phi}}\{\mathbf{X}\},
\label{equ_encoderMLP}
\end{equation}
where the weights and biases of encoder MLP are denoted as $\boldsymbol{\phi}$. In other words, the $\operatorname{EncoderMLP}_{\boldsymbol{\phi}}$ estimates the parameters of $q(\mathbf{z})$ from the input $\mathbf{X}$.
\par Similarly, from Equ. (\ref{equ_vaeloss_logpxz}), the VAE decoder becomes clear
\begin{equation}
(\boldsymbol{\mu}_p, \log\boldsymbol{\sigma}_p^2)=\operatorname{DecoderMLP}_{\boldsymbol{\theta}}\{\mathbf{z}\},
\label{equ_decoderMLP}
\end{equation}
where the weights and biases of decoder MLP are denoted as $\boldsymbol{\theta}$. In other words, the $\operatorname{DecoderMLP}_{\boldsymbol{\theta}}$ estimates the parameters of $p(\mathbf{X}|\mathbf{z})$ from the $\mathbf{z}$ sampled from Equ. (\ref{equ_reparatrick}).
\par Then the VAE architecture shown in Fig. \ref{fig_basicAE} (b) becomes clear by combining $\operatorname{EncoderMLP}_{\boldsymbol{\phi}}$ and $\operatorname{DecoderMLP}_{\boldsymbol{\theta}}$ together, where these two parts are bridged through the sampling of $\mathbf{z}$. And the VAE training objective is to minimize the loss function with respect to the network parameters $(\boldsymbol{\phi}, \boldsymbol{\theta})$. According to Equ. (\ref{equ_vaeloss}-\ref{equ_vaeloss_logpxz}), the overall VAE loss function is
\begin{align*}
\mathcal{L}_{\text{VAE}}&=\operatorname{KLD}\{q(\mathbf{z})||p(\mathbf{z})\} - \mathbb{E}_{q}[\log{p(\mathbf{X}}|\mathbf{z})]\\
&=\frac{1}{2}\sum_{n=1}^{N}\sum_{k=1}^{K}\{\frac{(\mathbf{x}_n[k]-\boldsymbol{\mu}_p[k])^2}{\boldsymbol{\sigma}_p[k]^2} + \log\boldsymbol{\sigma}_p[k]^2\}-\\
&\frac{1}{2}\sum_{d=1}^{D} \{1 + \log\boldsymbol{\sigma}_q[d]^2-\boldsymbol{\mu}_q[d]^2-\boldsymbol{\sigma}_q[d]^2\}.
\numberthis
\label{equ_vaeloss_overall}
\end{align*}

\subsection{Recurrent Autoencoder}
\par While we use the VI approach to learn the radar point cloud distribution at each frame, we also need a sequence-to-sequence modeling approach to learn distribution changes over multiple frames, as stated in Section \ref{sec_radarpointcloudmodel} previously.
\par The recurrent neural network (RNN) is such a basic sequence-to-sequence model for temporal applications. At every frame $l$, an RNN accepts two inputs, input from the sequence at the $l$-th frame $x_l$ and its previous hidden state ${h_{l-1}}$, to output a new hidden state $h_l$, calculated as:
\begin{equation}
h_l = \tanh(W*h_{l-1}+U*x_l) \textrm{     } \forall l=1,2, \dots,L
\end{equation}
where $W$ and $U$ are learnable weights (including the bias term, omitted for brevity), and $L$ is the length of the sequence. Note that at $l$=1, $h_0$ is defined as the initial RNN state that is either initialized as zeros, or randomly initialized. Also, note that the hidden state $h_{l}$ acts as an accumulated memory state as it continuously computed and updated with new information in the sequence. Based on the basic RNN, the Long-Short-Term-Memory (LSTM) and Gated-Recurrent-Units (GRUs) \cite{hochreiter1997long, cho2014learning} has been developed to solve the vanishing/exploding gradients issue in modeling long term dependencies \cite{bengio} in RNN. However, in our case, as a fall motion may last for about one second, that is ten frames for the radar data rate of ten frames per second, the long term dependency is not an issue here. Only the basic RNN is used for light computation load consideration.
\begin{figure}[H]
	\centering
	\input{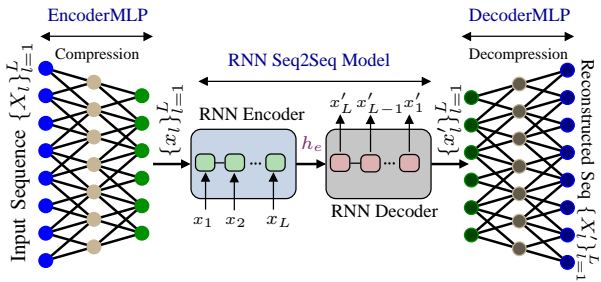}
	\caption{A depiction of a Recurrent Autoencoder (RAE). The input sequence $\{\mathbf{X}_l\}^L_{l=1}$ is first compressed to a embedded feature sequence $\{x_l\}^L_{l=1}$ on per frame basis through the EncoderMLP. An RNN Encoder iteratively processes the data over $L$ frames and the final hidden state $h_e$ is passed on to the RNN Decoder that outputs the reconstructed embedded feature sequence $\{x^\prime_l\}^L_{l=1}$ in reverse. Finally, $\{x^\prime_l\}^L_{l=1}$ are decompressed to reconstruct the sequence $\{\mathbf{X}^\prime_l\}^L_{l=1}$ through the DecoderMLP. The output sequence $\{\mathbf{X}^\prime_l\}^L_{l=1}$ is compared with the input sequence $\{\mathbf{X}_l\}^L_{l=1}$ to compute the reconstruction loss, which is desired to be low for an autoencoder.}
	\vspace{-0.2cm}
	\label{fig_RAE}
\end{figure}
\par The RNN-based autoencoder \cite{ref_RAE1} \cite{ref_RAE2}, or RAE as shown in Fig. \ref{fig_RAE}, is built upon the vanilla autoencoder architecture in Fig. \ref{fig_basicAE} (a). As the input is a time sequence of feature vectors, it has two dimensions, i.e., feature dimension and time dimension. In RAE, the EncoderMLP/DecoderMLP is for compressing and reconstructing the feature vector on per frame basis, and the RNN-Encoder/Decoder is for compressing and reconstructing the time sequence over multiple frames. Overall, the RAE reduces redundancy in both feature and time dimension.   
\begin{figure*}[ht]
	\centering
	\scalebox{0.9}{
		\input{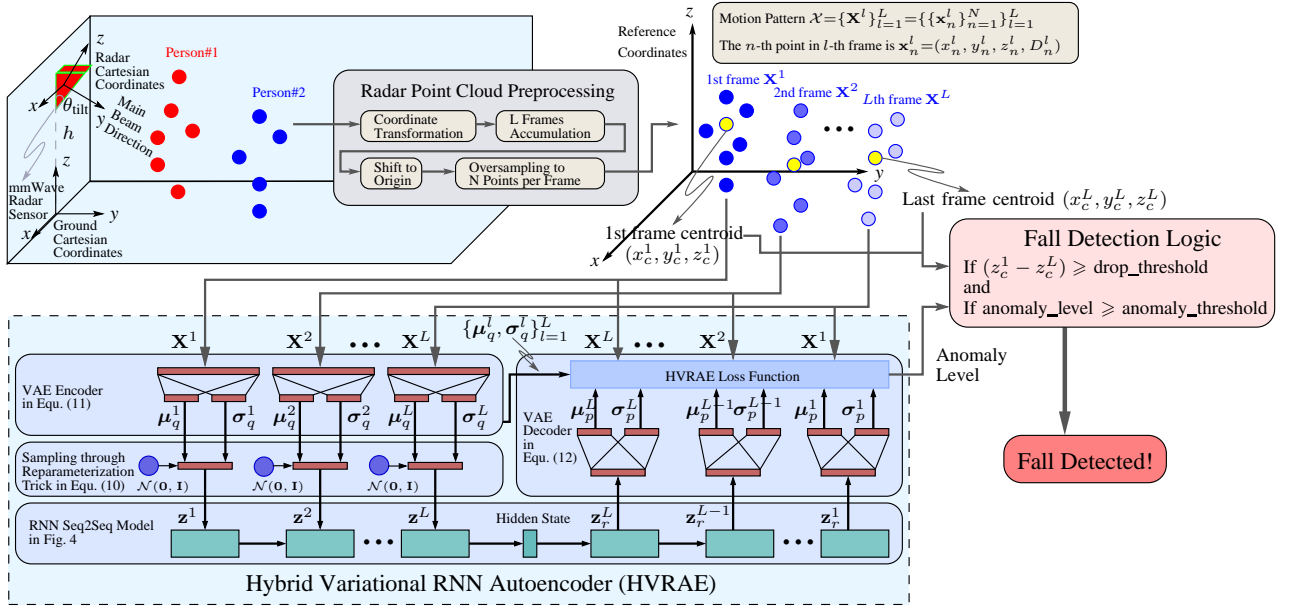}}
	\caption{An overview of the proposed \textit{mmFall} System. At each frame, we obtain the point cloud of a human body along with its centroid from mmWave radar sensor. After the preprocessing stage, we get a motion pattern $\mathcal{X}{=}\{\mathbf{X}^l\}^L_{l=1}$ in the reference coordinates. For each $l$-th frame, we use the VAE Encoder to model the mean $\boldsymbol{\mu}^l_q$ and variance $\boldsymbol{\sigma}^l_q$ of the factorized Gaussian family $q(\mathbf{z}^l){=}\prod_{d=1}^D\mathcal{N}(z^l_d|\mu^l_d, \sigma^l_d)$ that aims to approximate the true posterior $p(\mathbf{z}^l|\mathbf{X}^l)$ of the latent motion state $\mathbf{z}^l$, where $D$ is the predetermined length of $\mathbf{z}$. Then we use the reparameterization trick to sample $\mathbf{z}^l$ from $q(\mathbf{z}^l)$. After we have a sequence of latent motion states $\mathcal{Z}=\{\mathbf{z}^1,...,\mathbf{z}^L\}$, we use the RAE to compress and then reconstruct it as $\mathcal{Z}_r=\{\mathbf{z}^1_r,...,\mathbf{z}^L_r\}$. Based on $\mathcal{Z}_r$, we use the VAE Decoder to model the mean $\boldsymbol{\mu}^l_p$ and variance $\boldsymbol{\sigma}^l_p$ of the likelihood $p(\mathbf{X}^l|\mathbf{z}^l)$. With $(\boldsymbol{\mu}_q, \boldsymbol{\sigma}_q)$, $(\boldsymbol{\mu}_p, \boldsymbol{\sigma}_p)$ and $\mathcal{X}$, we are able to compute the HVRAE loss defined in Equ. \ref{equ_vraeloss} as an indication of anomaly level. In the fall detection logic, if a sudden drop of centroid height is detected at the same time when the HVRAE outputs an anomaly spike, we claim a fall detection.}
	\vspace*{-0.2cm}
	\label{fig_proposedsys}
\end{figure*}
\section{Proposed System}\label{sec_proposedsystem}
\par To effectively learn the motion pattern of human body, which is formed by a sequence of radar point cloud, for fall detection in a semi-supervised approach, we propose a Hybrid Variational RNN AutoEncoder (HVRAE) which has two autoencoder substructures, i.e., VAE for learning radar point cloud distribution on per frame basis and RAE for learning the change of distribution over multiple frames. The HVRAE is trained only on normal ADL, such that an `unseen' fall will cause a spike in the loss or anomaly level. If the height of body centroid, which is estimated from the point cloud, drops suddenly at the same time, a fall is detected. The proposed system, called \textit{mmFall}, including both hardware and software, is presented in Fig. \ref{fig_proposedsys}.
\subsection{Data Preprocessing}
\par With a proper mmWave radar sensor, we are able to collect the radar point cloud, as shown in Fig. \ref{fig_radarsummary} (b). In Fig. \ref{fig_proposedsys}, the radar sensor could be mounted on the wall in a room with a height of $h$ over the head of people, and could also be rotated with an angle $\theta_{\text{tilt}}$ so that it has a better coverage of the room. The radar sensor can detect multiple moving persons simultaneously, each person has a unique target ID as a result of the clustering/tracking algorithms. With the multiple frame data with the same target ID, we can analyze the motion of the person associated this target ID. In other words, each person's motion analysis can be processed separately based on the target ID. Afterwards, we will only discuss the single-person scenario for brevity. \par We then propose a data preprocessing flow denoted in Fig. \ref{fig_proposedsys} for the following reasons.
\par The original measurement for each point in the radar point cloud is in the radar spherical coordinates. We need to transfer it to the radar Cartesian coordinates, and then to the ground Cartesian coordinates on the basis of the tilt angle and height. Therefore, we have a transformation matrix as
\begin{align*}
\Big[\begin{smallmatrix}
x\\
y\\
z
\end{smallmatrix}\Big]=\Big[\begin{smallmatrix}
1 & 0 & 0\\
0 & \cos\theta_{\text{tilt}} & \sin\theta_{\text{tilt}}\\
0 & -\sin\theta_{\text{tilt}} & \cos\theta_{\text{tilt}}\\
\end{smallmatrix}\Big]\Big[\begin{smallmatrix}
r\cos\theta_{\text{EL}}\sin\theta_{\text{AZ}}\\
r\cos\theta_{\text{EL}}\cos\theta_{\text{AZ}}\\
r\sin\theta_{\text{EL}}\\
\end{smallmatrix}\Big]+\Big[\begin{smallmatrix}
0\\
0\\
h
\end{smallmatrix}\Big],
\numberthis
\label{equ_coordtrans}
\end{align*}
where $(r, \theta_{\text{AZ}}, \theta_{\text{EL}})$ is range, azimuth angle and elevation angle in the radar spherical coordinates, $\theta_{\text{tilt}}$ is radar tilt angle, $h$ is the radar platform height, and $[x, y, z]^T$ is the result in the ground Cartesian coordinates.
\par After coordinate transformation, at each frame we obtain a radar point cloud, in which each point is a vector of $(x, y, z, {DP})$ where ${DP}$ is the Doppler from the original radar measurement. And we also have the centroid $(x_c, y_c, z_c)$ as a result of the clustering/tracking algorithms in the radar.
\par We accumulate the current frame's previous $L$ frames including itself as a motion pattern. The value of $L$ equals to the radar frame rate in frames per second (fps) multiplied by the predetermined detection window in seconds. For each motion pattern with $L$ frames, we subtract the $x$ and $y$ value of each point in each frame from the $x_c$ and $y_c$ value of centroid in the first frame, respectively. In this way, we shift the motion pattern to the origin of a reference coordinates.  
\begin{algorithm}[htp]
	\caption{Data Oversampling Method}
	\label{alg1}
	\KwIn{Input dataset $\mathbf{X}=\{\mathbf{x}_i\}^{M}_{i=1}$ with a length of ${M}$, $M$ is a random number, each data sample $\mathbf{x}_i$ is a vector. $N$, target length after oversampling. $N$ is always $\geqslant M$.}
	\KwOut{$\mathbf{X}^\prime=\{\mathbf{x}^\prime_i\}^{N}_{i=1}$ with a length of ${N}$.}
	\BlankLine
	$\hat{\boldsymbol{\mu}}=\frac{1}{M}\sum_{i=1}^{M}\mathbf{x}_i$ \tcp*[h]{Get the estimated mean}\\
	\For{$i=1$ to $N$} {
		\eIf(\tcp*[h]{Rescale and shift}){$i \leqslant M$}{
			$\mathbf{x}^\prime_i = \sqrt{\frac{N}{M}}\mathbf{x}_i + \hat{\boldsymbol{\mu}} - \sqrt{\frac{N}{M}}\hat{\boldsymbol{\mu}}$; \label{alg1_step4}
		} 
		(\tcp*[h]{Pad with $\hat{\boldsymbol{\mu}}$}){
			$\mathbf{x}^\prime_i = \hat{\boldsymbol{\mu}}$; \label{alg1_step6}
		}		
	}
\end{algorithm}  
\par At each frame, the number of points in the radar point cloud is random due to the nature of radar measurement. We need a data oversampling method to meet the fixed input of the HVRAE model. The traditional oversampling method in deep learning is such as zero-padding or random oversampling. Zero-padding simply adds more zeros into the original data and random sampling simply duplicates some original data. Using both these two oversampling methods, the distribution of the input may be changed. However, our purpose is to learning the distribution of radar point cloud and changing the distribution is definitely not what we want. Therefore, we propose a novel data oversampling Algorithm \ref{alg1} that extends the original point cloud to a fixed number while keeping its distribution (mean and covariance) the same. The proof of this algorithm is in Appendix \ref{algo1_proof}.
\par Finally, we obtain a motion pattern $\mathcal{X}$ in the reference coordinates,
\begin{align*}
\mathcal{X}{=}\{\mathbf{X}^l\}^L_{l=1}&{=}\{\{\mathbf{x}^l_n\}^N_{n=1}\}^L_{l=1}\\
&{=}\{\{(x^l_n, y^l_n, z^l_n, {DP}^l_n)\}^N_{n=1}\}^L_{l=1},
\numberthis
\label{equ_motionpattern}
\end{align*}
where $L$ is the number of frames in the motion pattern; $N$ is the number of points at each frame; $\mathbf{X}^l$ is the $l$-th frame point cloud; $\mathbf{x}^l_n$ is the $n$-th point in $l$-th frame, that is also a 4D vector of $(x^l_n, y^l_n, z^l_n, {DP}^l_n)$. We also have the centroid $\{(x^l_c, y^l_c, z^l_c)\}^L_{l=1}$ over $L$ frames. Afterwards, we use the superscript $l$ to denote the frame index.

\subsection{HVRAE Model}
\par The HVRAE architecture is shown in Fig. \ref{fig_proposedsys} and detailed in the caption. The HVRAE model is a combination of VAE and RAE, discussed in the previous section. The HVRAE loss $\mathcal{L}_{\text{HVRAE}}$ is the VAE loss $\mathcal{L}_{\text{VAE}}$ in Equ. (\ref{equ_vaeloss_overall}) over all the $L$ frames. Then, we have
\begin{align*}
\mathcal{L}_{\text{HVRAE}}&=\sum_{l=1}^{L}\Big\{\operatorname{KLD}\{q(\mathbf{z}^l)||p(\mathbf{z}^l)\} - \mathbb{E}_{q}[\log p(\mathbf{X}^l|\mathbf{z}^l)]\Big\}\\
&=\sum_{l=1}^{L}\Big\{ \frac{1}{2}\sum_{n=1}^{N}\sum_{k=1}^{K}\{\frac{(\mathbf{x}^l_n[k]-\boldsymbol{\mu}^l_p[k])^2}{\boldsymbol{\sigma}^l_p[k]^2} + \log\boldsymbol{\sigma}^l_p[k]^2\}\\
&- \frac{1}{2}\sum_{d=1}^{D} \{1 + \log\boldsymbol{\sigma}^l_q[d]^2 - \boldsymbol{\mu}^l_q[d]^2 - \boldsymbol{\sigma}^l_q[d]^2\}\Big\}
\numberthis
\label{equ_vraeloss}
\end{align*}
where $L$, $N$ and $\mathbf{x}^l_n$ are from the motion pattern in Equ (\ref{equ_motionpattern}); $K$ is the length of point vector, in our case $K{=}4$ as each point is a 4D vector; $D$ is the length of latent motion state $\mathbf{z}$; $(\boldsymbol{\mu}_q, \boldsymbol{\sigma}_q)$ and $(\boldsymbol{\mu}_p, \boldsymbol{\sigma}_p)$ are parameters of factorized Gaussian $q(\mathbf{z})$ and likelihood $p(\mathbf{X}|\mathbf{z})$, respectively, both are modeled through the architecture in Fig. \ref{fig_proposedsys}.
\par For HVRAE training, the objective is to minimize $\mathcal{L}_{\text{HVRAE}}$ with respect to the network parameters. The standard stochastic gradient descent algorithm Adam \cite{ref_vae2} is used.
\par It is noted that, for the implementation of VAE Encoder/Decoder in HVRAE, only a dense layer or fully-connected layer is used, as the model should be invariant to the order of point cloud at each frame.
\subsection{Fall Detection Logic}
\par In a semi-supervised learning approach, we train this HVRAE model only on normal ADL, which are easy to collect compared to falls. For normal ADL, the HVRAE will output a low $\mathcal{L}_{\text{HVRAE}}$ as this is the training objective. In the inference stage, the model will generate a high loss $\mathcal{L}_{\text{HVRAE}}$ when an `unseen' motion happens, such as fall occurs. Therefore, we denote the HVRAE loss $\mathcal{L}_{\text{HVRAE}}$ as an anomaly level measure of human body motion.
\par Along with the body centroid height $\{z^l_c\}^L_{l=1}$ over $L$ frames, we can calculate the drop of centroid height as $z^1_c-z^L_c$ during this motion. Then we propose a fall detection logic as in Fig. \ref{fig_proposedsys}, that is if the centroid height drop is greater than a threshold at the same time when the anomaly level is greater than a threshold, we claim a fall detection.
\par According to the fall definition from WHO as in the Section I, in the proposed \textit{mmFall} system, the HVRAE measures the inadvertence or anomaly level of the motion, while the centroid height drop indicates the motion of coming to rest on a lower level.

\section{Experimental Results and Discussion}\label{sec_experiment}
\begin{figure*}[ht]
	\centering
	\includegraphics[scale=0.135]{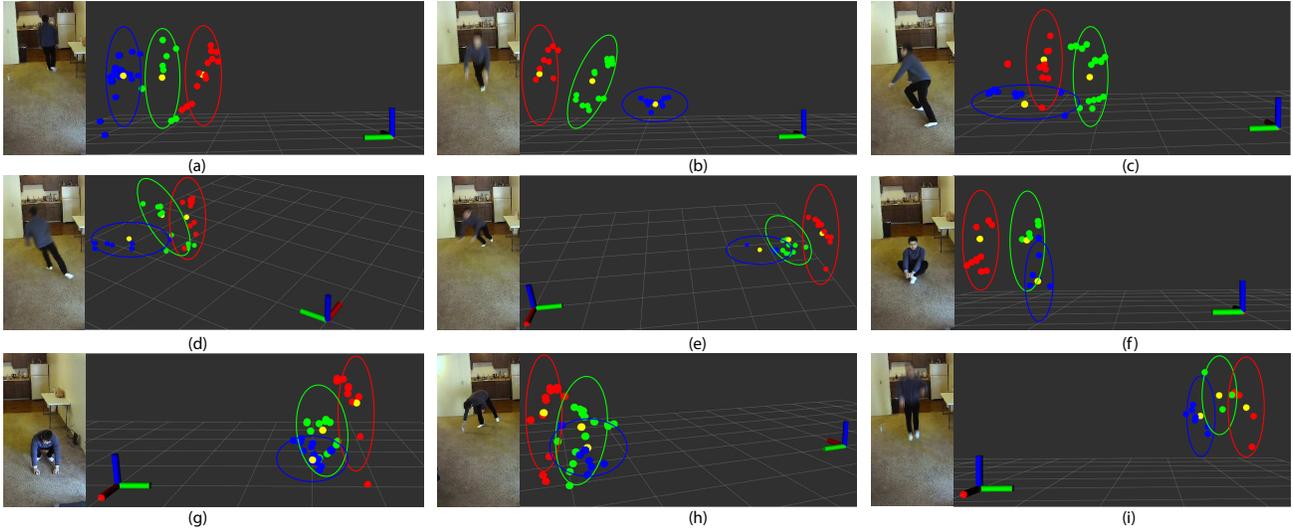}
	\vspace*{-0.1cm}
	\caption{Motion patterns in dataset $DS_1$ along with the associated camera view. Only the ellipse was manually added for depicting the distribution of point cloud. For the points, different color indicates the frame in time order: red, green, and then blue, while the yellow point indicates the centroid estimated by the mmWave radar sensor. For simplicity, we showed the frames with the increment of five frames. Each frame is 0.1 seconds. Please compare this figure with Fig. \ref{fig_depiction_motionpattern}. For the coordinates, red is the cross-radar direction, green is the forward direction, and blue is the height direction. (a) Randomly walking; (b) Forward fall; (c) Backward fall; (d) Left fall; (e) Right Fall; (f) Sitting down on the floor; (g) Crouching; (h) Bending; (i) jump.}
	\vspace*{-0.2cm}
	\label{fig_motionpattern}
\end{figure*}
To verify the effectiveness of the proposed system, we used a mmWave radar sensor to collect experimental data and implemented the proposed \textit{mmFall} system along with two baselines for performance evaluation and comparison. 
\subsection{Hardware Configuration and Experiment Setup}
\par We adopt the Texas Instrument (TI) AWR1843BOOST mmWave FMCW radar evaluation board \cite{ref_AWR1843BOOST} for radar point cloud acquisition. This radar sensor has three transmitting antenna channels and four receiving antenna channels, as shown in Fig. \ref{fig_radarsummary} (a). The middle transmitting channel is displaced above the other two by a distance of half a wavelength. Through the direction-of-angle (DOA) algorithm using multiple-input and multiple-output (MIMO), it can achieve 2x4 MIMO in azimuth and 2x1 MIMO in elevation. Thus, we have 3D positional measurement of each point. Plus the 1D Doppler, we finally have a 4D radar point cloud. Based on a demo project from TI \cite{ref_TIDEMOPROJ}, we configure the radar sensor with the parameters listed in Table \ref{tab_radarparam}.
\begin{table}[h]
	\renewcommand{\arraystretch}{1.3}
	\caption{mmWave FMCW radar parameter configuration. Refer to \cite{ref_AWRWAVEFORM} for waveform details. $f_s$, FMCW starting frequency. $\text{BW}$, FMCW bandwidth. $r_{\text{Chirp}}$, FMCW chirp rate. $f_{\text{ADC}}$, ADC sampling rate. $N_{\text{Fast}}$, ADC samples per chirp. CPI, coherent processing interval. $N_{\text{Slow}}$, chirps per CPI per transmitting channel. $T_{\text{Frame}}$, duration of one frame. $\Delta R$, range resolution. $R_{\text{max}}$, maximum unambiguous range. $\Delta D$, Doppler resolution. $D_{\text{max}}$, maximum unambiguous Doppler. $\Delta\theta_{\text{AZ}}$, azimuth angle resolution. $\Delta\theta_{\text{EL}}$, elevation angle resolution. $r_{\text{Frame}}$, frame rate in frames per second.}
	\label{tab_radarparam}
	\centering
	\resizebox{\columnwidth}{!}{%
	\begin{tabular}{|c||c||c|c||c||c|}
		\hline
		\bfseries Parameter & \bfseries Value & \bfseries Unit & \bfseries Parameter & \bfseries Value & \bfseries Unit \\
		\hline
		$f_s$ & $77$ & GHz & $\Delta R$ & $0.078$ & m\\
		\hline
		$\text{BW}$ & $1.92$ & GHz & $R_{\text{max}}$ & $9.99$ & m\\
		\hline
		$r_{\text{Chirp}}$ & $30$ & MHz/us & $\Delta D$ & $0.079$ & m/s\\
		\hline
		$f_{\text{ADC}}$ & $2$ & MHz &  $D_{\text{max}}$ & $\pm2.542$ & m/s\\
		\hline
		$N_{\text{Fast}}$ & $128$ & & MIMO & 2x4/2x1 & AZ/EL\\
		\hline
		CPI & $24.2$ & ms & $\Delta\theta_{\text{AZ}}$ & $15$ & deg\\    
		\hline
		$N_{\text{Slow}}$ & $64$ & & $\Delta\theta_{\text{EL}}$ & $57$ & deg\\
		\hline
		$T_{\text{Frame}}$ & $100$ & ms & $r_{\text{Frame}}$ & $10$ & fps\\
		\hline
	\end{tabular}
	}
	\vspace*{-0.2cm}
\end{table}
\par Based on the Robotic Operating System (ROS) on an Ubuntu laptop, we developed an interface program to connect the TI AWR1843BOOST and collect the radar point cloud over the USB port. Then we set up the equipment in the living room (2.7m*8.2m*2.7m) in an apartment, as shown in Fig. \ref{fig_radarsummary} (a). There are two large desks in the living room and most area is relatively empty. More occlusion discussion can be found in Section \ref{sec_discussion}. The radar sensor was put on top of a tripod with a height of 2 meters, and rotated with a tilt angle of 10 degrees for better area coverage. Later, we processed the collected data offline using a Jupyter Notebook that you can found in the GitHub repository.

\subsection{Data Collection}
\par During the experiment, the first two authors, as shown in Fig. \ref{fig_radarsummary} (b), collected three datasets in Table \ref{tab_collecteddataset} together. Firstly, we collected the $DS_0$ dataset which contains about two hours of normal ADL without any labels for training. Secondly, in the $DS_1$ dataset, we collected randomly walking along with one sample of each other motion, including fall, etc. We showed the motion pattern for every motion in Fig. \ref{fig_motionpattern} for visualization purposes. Lastly, we collected a comprehensive inference dataset $DS_2$ and manually labeled the frame index when a fall happens as the ground truth, and it is used for overall inference performance evaluation. It is noted that in $DS_1$ and $DS_2$, both the fall and jump are anomalies that can not be found in $DS_0$. We expect that HVRAE will output an anomaly level spike for both fall and jump, but the fall detection logic involving the centroid height drop will reject the jump but detect the fall.
\begin{table}[h]
	\renewcommand{\arraystretch}{1.3}
	\vspace*{-0.0cm}
	\caption{Collected Dataset.}
	\label{tab_collecteddataset}
	\centering
	\resizebox{\columnwidth}{!}{%
	\begin{tabular}{|c||l|}
		\hline
		\bfseries Name & \makecell{\bfseries Description}  \\
		\hline
		 \multirow{2}{*}{$DS_{\text{0}}$} & Two hours of normal ADL, including randomly walking, sitting\\ 
		 & on the floor, crouching, bending, etc. No labeling. \\
		 \hline
		 \multirow{3}{*}{$DS_{\text{1}}$} & Randomly walking with one forward fall, one backward fall, \\
		 & one left fall, one right fall, one sitting on the floor, one \\
		 &  crouching, one bending, and one jump. \\
		 \hline
		 \multirow{3}{*}{$DS_{\text{2}}$} & Randomly walking with 15 forward falls, 15 backward falls, 10\\
		 & left falls, 10 right falls, 50 sitting on the floor, 50 crouching,\\
		 & 50 bending, and 50 jump. Labeling fall as ground truth.\\
		\hline
	\end{tabular}
	}
	\vspace*{-0.4cm}
\end{table}
\subsection{Model Implementation and Two Baselines}
\begin{figure*}[ht]
	\centering
	\includegraphics[scale=0.4]{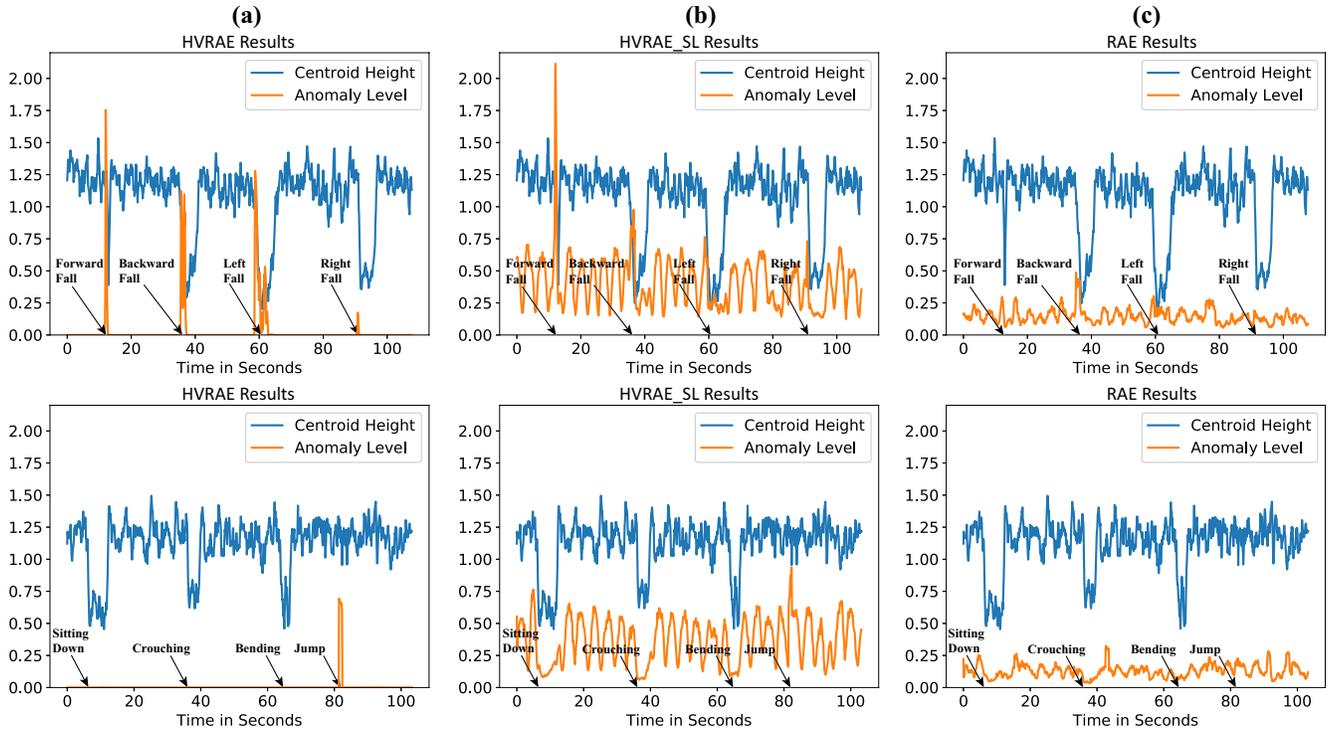}
	\caption{Inference results of the models listed in Tab \ref{tab_models} on the dataset $DS_1$ described in Tab \ref{tab_collecteddataset}. In each figure, the blue line represents the body's centroid height, and the orange line represents the model's loss output, or anomaly level. Only the black text and arrows were manually added as the ground truth when a motion happens. Except for the motion indicated by the black text, the rest of time are always randomly walking. (a) HVRAE inference results: The HVRAE model can clearly generate a spike in anomaly level when fall/jump happens while keeping low anomaly level for normal motions. Jump is another abnormal motion that does not appear in the training dataset $DS_0$, but the fall detection logic involving the body centroid drop at the same time will reject jump. On the other hand, without the help of anomaly level it is difficult to distinguish fall from other motions if only the change of centroid height is considered; (b) HVRAE\_SL inference results: The HVRAE\_SL can also have anomaly level spike generation for fall/jump but suffer significant noise during normal motion occurrence. For example, the `Sitting Down' and the `Right Fall' have almost the same anomaly level output. As a result, either the `Sitting Down' causes a false alarm, or `Right Fall' causes a missed detection, depending on the threshold; (c) Vanilla RAE inference results: The vanilla RAE model can not effectively learn the anomaly level for `unseen' motions.}
	\vspace*{-0.2cm}
	\label{fig_D1results}
\end{figure*}
\par We first implemented the proposed \textit{mmFall} system in Fig. \ref{fig_proposedsys} on Keras (Tensorflow backend), with loss function in Equ. (\ref{equ_vraeloss}). In this implementation, we set the number of frames, $L$, equal to 10 for a one-second detection window with 10 fps radar data rate; the number of points each frame $N$ equal to 64 for data oversampling. Thus, the motion pattern $\mathcal{X}$, i.e., the model input, is 10*64*4. We set the length of latent motion state $z$, $D$, equal to 16. For performance comparison purposes, we also implemented two other baselines. All the three models are listed in Table \ref{tab_models}.
\par The baseline HVRAE\_SL is the same as the proposed \textit{mmFall} system except for using a simplified loss function in Equ. (\ref{equ_vrae_sl_loss}). The simplified loss function Equ. (\ref{equ_vrae_sl_loss}) is based on a weak assumption on likelihood, that is $p(\mathbf{X}|\mathbf{z})$ follows a Gaussian with identity covariance, i.e., $\mathcal{N}(\boldsymbol{\mu}_p, \mathbf{I})$. This leads to that the $\boldsymbol{\sigma}_p$ term in Equ. (\ref{equ_vraeloss}) is ignored, or
\begin{align*}
\mathcal{L}_{\text{HVRAE\_SL}}&{=}\sum_{l=1}^{L}\Big\{ \frac{1}{2}\sum_{n=1}^{N}\sum_{k=1}^{K}\{(\mathbf{x}^l_n[k]-\boldsymbol{\mu}^l_p[k])^2\}-\frac{1}{2}*\\
&\sum_{d=1}^{D} \{1 + \log\boldsymbol{\sigma}^l_q[d]^2 - \boldsymbol{\mu}^l_q[d]^2 - \boldsymbol{\sigma}^l_q[d]^2\}\Big\}.
\numberthis
\label{equ_vrae_sl_loss}
\end{align*}
To compare HVRAE with HVRAE\_SL, we will verify that the concept that the covariance represents the pose contributes to the radar point cloud learning for human motion inference, as discussed in Section \ref{sec_radarpointcloudmodel}.
\par Another baseline is RAE with MSE loss in Fig. \ref{fig_RAE}, which uses MLP in the feature dimension instead of VI approach in HVRAE every frame. To compare HVRAE with RAE, we will show that the VI approach for motion state inference based on the distribution of radar point cloud makes more sense than the vanilla MLP feature compression.
\begin{table}[h]
	\renewcommand{\arraystretch}{1.3}
	\caption{Implemented Models.}
	\label{tab_models}
	\centering
	\resizebox{\columnwidth}{!}{%
	\begin{tabular}{|c||l|}
		\hline
		\bfseries Name & \makecell{\bfseries Description}  \\
		\hline
		\multirow{2}{*}{HVRAE} & The proposed HVRAE and fall detection logic\\ 
		& in Fig. \ref{fig_proposedsys} with loss function in Equ. (\ref{equ_vraeloss}). \\
		\hline
		\multirow{2}{*}{HVRAE\_SL} & The proposed HVRAE and fall detection logic\\ 
		& in Fig. \ref{fig_proposedsys} with simplified loss in Equ. (\ref{equ_vrae_sl_loss}). \\
		\hline
		\multirow{2}{*}{RAE} & The vanilla RAE in Fig. \ref{fig_RAE} with MSE loss function\\
		& and fall detection logic in Fig. \ref{fig_proposedsys}. \\
		\hline
	\end{tabular}
	}
	\vspace*{-0.2cm}
\end{table}
\subsection{Training and Inference}
\par First, we trained these three models on the normal dataset $DS_0$, and then tested on dataset $DS_1$ in which there are some normal motions as in $DS_0$ and two different `unseen' motions, i.e., fall and jump, that do not appear in $DS_0$. The anomaly level outputted by these three models on $DS_1$ is shown in Fig. \ref{fig_D1results}. The proposed HVRAE model can generate significant anomaly level for fall and jump while keeping low for normal motions. Along with the fall detection logic involving body centroid drop, the jump will be rejected, and only fall will be detected. As a comparison, the HVRAE\_SL model suffers great noise during normal motions that easily leads to false alarm, and the vanilla RAE model can not learn the anomaly level effectively.
\par
\begin{figure}[H]
	\centering
	\includegraphics[scale=0.5]{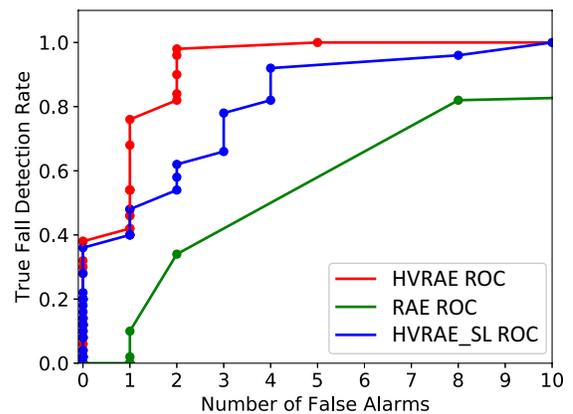}
	\caption{ROC curves for all the three models.}
	\label{fig_ROC}
\end{figure}

 Finally, we tested these three well-trained models on the dataset $DS_2$. In $DS_2$, there are 50 falls with manually labeled `ground truth fall frame index' when a fall happens, along with many other different motions without labeling. The fall detection logic will detect the frame index when a fall happens. We allow a flexible detection, i.e., if the `detected fall frame index' falls into the 1-second detection window centered at one `ground truth fall frame index', we treat it as true positive. In this experiment, for the fall detection logic we fixed the threshold of centroid height drop as 0.6 meters. By varying the anomaly level threshold, we got the Receiver Operating Characteristic (ROC) curves as shown in Fig. \ref{fig_ROC}. The false alarm will cause waste of caring resources, and high true fall detection rate guarantees the elderly safety. Therefore, we want to achieve as high fall detection rate as possible at the expense of a few false alarms. From the ROC, we clearly see that HVRAE outperformed the other two baselines. Specifically, at the expense of two alarms, our HVRAE model can achieve 98\% fall detection rate out of 50 falls, while the HVRAE\_SL can only achieve around 60\% and the vanilla RAE can only achieve around 38\%.

\subsection{Limitations of Current Research and Future Work}\label{sec_discussion}
\par In this research, we did the experiment in a relatively empty apartment where occlusion is not a problem. To have more comprehensive results, an experiment in a complex living environment is necessary. Also, the radar sensor is essentially robust to occlusion due to the nature of radio frequency (RF) signal. Basically, the Signal-to-Noise Ratio (SNR), that is related to the radar hardware's noise figure and transmitting power, determines the occlusion performance. A powerful radar sensor can even `see' through the wall \cite{ref_MITRFCAPTURE}. To obtain a more practical validation, in the future we aim to incorporate necessary hardware engineering to improve the SNR of the current radar sensor, and demonstrate the performance in an apartment with lots of furniture. 

\par The human subjects in the experiment are the first two authors who have a very similar body figure ($\sim175cm/75kg$). As we view the point cloud of human body as a distribution, typically, if there is a huge difference of body figure than the subjects' in this experiment, the distribution should also be quite different. Therefore, the model trained in this experiment can not be directly applied to the person with a significantly different body figure, for example $\sim190cm/110kg$. In the future, we will collect more training data from multiple human subjects with a wide range of build/height, to make the model be able to cover more cases.

\section{Conclusion}\label{sec_conclusion}
In this study, we used a mmWave radar sensor for fall detection on the basis of its advantages such as privacy-compliant, non-wearable, sensitive to motions, etc. We made an assumption that the radar point cloud for the human body can be viewed as a multivariate Gaussian distribution, and the distribution change over multiple frames has a unique pattern for different motions. And then, we proposed a Hybrid Variational RNN AutoEncoder to effectively learn the anomaly level of `unseen' motion, such as fall, that does not appear in the normal training dataset. We also involved a fall detection logic that checks the body centroid drop to further confirm the anomaly motion is fall. In this way, we detected the fall in a semi-supervised learning approach that does not require the difficult fall data collection and labeling. The experiment results showed our proposed system can achieve 98\% detection rate out of 50 falls at the expense of just two false alarms, and outperformed the other two baselines. In the future, we will have necessary hardware engineering to improve the SNR and demonstrate the occlusion performance of the mmWave radar sensor in a complex ling environment and also collect more training data from people with different body figures.

\appendices
\section{Proof of Equ. \ref{equ_vaeloss_KLD}}
\label{equ8_proof}
\begin{align*}
&\operatorname{KLD}\{q(\mathbf{z})||p(\mathbf{z})\}\\
=&\operatorname{KLD}\Bigg\{\prod_{d=1}^D\mathcal{N}(\mathbf{z}[d]|\boldsymbol{\mu}_q[d], \boldsymbol{\sigma}_q[d])||\prod_{d=1}^D\mathcal{N}(\mathbf{z}[d]|0, 1)\Bigg\}\\
:=&\int...\int \Bigg\{\prod_{d=1}^D\mathcal{N}(\mathbf{z}[d]|\boldsymbol{\mu}_q[d], \boldsymbol{\sigma}_q[d]) *\\
&\log\frac{\prod_{d=1}^D\mathcal{N}(\mathbf{z}[d]|\boldsymbol{\mu}_q[d], \boldsymbol{\sigma}_q[d])}{\prod_{d=1}^D\mathcal{N}(\mathbf{z}[d]|0, 1)}\Bigg\} d\mathbf{z}[1]...d\mathbf{z}[D]\\
=&\int...\int \Bigg\{\prod_{d=1}^D\mathcal{N}(\mathbf{z}[d]|\boldsymbol{\mu}_q[d], \boldsymbol{\sigma}_q[d]) *\\
&\sum_{d=1}^{D}\log\frac{\mathcal{N}(\mathbf{z}[d]|\boldsymbol{\mu}_q[d], \boldsymbol{\sigma}_q[d])}{\mathcal{N}(\mathbf{z}[d]|0, 1)}\Bigg\} d\mathbf{z}[1]...d\mathbf{z}[D]\\
=&\sum_{d=1}^{D} \Bigg\{\int...\int \prod_{d=1}^D\mathcal{N}(\mathbf{z}[d]|\boldsymbol{\mu}_q[d], \boldsymbol{\sigma}_q[d]) *\\
&\log\frac{\mathcal{N}(\mathbf{z}[d]|\boldsymbol{\mu}_q[d], \boldsymbol{\sigma}_q[d])}{\mathcal{N}(\mathbf{z}[d]|0, 1)} d\mathbf{z}[1]...d\mathbf{z}[D]\Bigg\}\\
=&\sum_{d=1}^{D} \Bigg\{\int \mathcal{N}(\mathbf{z}[d]|\boldsymbol{\mu}_q[d], \boldsymbol{\sigma}_q[d]) \log\frac{\mathcal{N}(\mathbf{z}[d]|\boldsymbol{\mu}_q[d], \boldsymbol{\sigma}_q[d])}{\mathcal{N}(\mathbf{z}[d]|0, 1)} d\mathbf{z}[d] \Bigg\}\\
=&\sum_{d=1}^{D} \Bigg\{\int \mathcal{N}(\mathbf{z}[d]|\boldsymbol{\mu}_q[d], \boldsymbol{\sigma}_q[d])\\ 
&\bigg\{\log\frac{1}{\sqrt{\boldsymbol{\sigma}_q[d]^2}} - \frac{(\mathbf{z}[d]-\boldsymbol{\mu}_q[d])^2}{2\boldsymbol{\sigma}_q[d]^2} + \frac{\mathbf{z}[d]^2}{2} \bigg\} d\mathbf{z}[d] \Bigg\}\\
=&-\frac{1}{2}\sum_{d=1}^{D} \Bigg\{ \int \mathcal{N}(\mathbf{z}[d]|\boldsymbol{\mu}_q[d], \boldsymbol{\sigma}_q[d])\log{\boldsymbol{\sigma}_q[d]^2}d\mathbf{z}[d] + \\ 
&\int \mathcal{N}(\mathbf{z}[d]|\boldsymbol{\mu}_q[d], \boldsymbol{\sigma}_q[d])\frac{(\mathbf{z}[d]-\boldsymbol{\mu}_q[d])^2}{\boldsymbol{\sigma}_q[d]^2}d\mathbf{z}[d] -\\
&\int \mathcal{N}(\mathbf{z}[d]|\boldsymbol{\mu}_q[d], \boldsymbol{\sigma}_q[d]){\mathbf{z}[d]^2}d\mathbf{z}[d] \Bigg\}\\
=&-\frac{1}{2}\sum_{d=1}^{D} \Bigg\{ \log{\boldsymbol{\sigma}_q[d]^2}*\int \mathcal{N}(\mathbf{z}[d]|\boldsymbol{\mu}_q[d], \boldsymbol{\sigma}_q[d])d\mathbf{z}[d] + \\ 
&\frac{1}{\boldsymbol{\sigma}_q[d]^2}*\int \mathcal{N}(\mathbf{z}[d]|\boldsymbol{\mu}_q[d], \boldsymbol{\sigma}_q[d]){(\mathbf{z}[d]-\boldsymbol{\mu}_q[d])^2}d\mathbf{z}[d] -\\
&\int \mathcal{N}(\mathbf{z}[d]|\boldsymbol{\mu}_q[d], \boldsymbol{\sigma}_q[d]){\mathbf{z}[d]^2}d\mathbf{z}[d] \Bigg\}\\
&(As \: \mathbb{E}[(x-\mu)^2]=\sigma^2 \: and \: \mathbb{E}[x^2]=\mu^2+\sigma^2, \: then)\\
=&-\frac{1}{2}\sum_{d=1}^{D} \Bigg\{1 + \log\boldsymbol{\sigma}_q[d]^2-\boldsymbol{\mu}_q[d]^2-\boldsymbol{\sigma}_q[d]^2 \Bigg\},
\numberthis
\label{equ_vaeloss_KLD_detailed}
\end{align*}
where $(\boldsymbol{\mu}_q, \boldsymbol{\sigma}_q)$ is the mean and variance of of the factorized Gaussian $q(\mathbf{z})$ with $D$-dimensional latent vector $\mathbf{z}$. 

\section{Proof of Algorithm 1}
\label{algo1_proof}
\par Given a set of statistically independent and identically distributed (i.i.d.) data $\mathbf{X}=\{\mathbf{x}_i\}^{M}_{i=1}$ drawn from a multivariate Gaussian random variable $\mathcal{N}(\boldsymbol{\mu}, \boldsymbol{\sigma})$, thus the maximum likelihood (ML) estimator of its mean $\boldsymbol{\mu}$ and covariance $\boldsymbol{\sigma}$ is,
\begin{align*}
\hat{\boldsymbol{\mu}}=\frac{1}{M}\sum_{i=1}^{M}\mathbf{x}_i, \ \hat{\boldsymbol{\sigma}}=\frac{1}{M}\sum_{i=1}^{M}\{\mathbf{x}_i-\hat{\boldsymbol{\mu}}\}^2.
\numberthis
\label{equ_inputmeanvariance}
\end{align*}
\par For the output dataset $\mathbf{X^\prime}=\{\mathbf{x^\prime}_i\}^{N}_{i=1}$, its first $M$ elements are modified from the input dataset according to Step \ref{alg1_step4} in Algorithm (\ref{alg1}), and its last $(N-M)$ elements are simply the mean of the input dataset according to Step \ref{alg1_step6} in Algorithm (\ref{alg1}). Thus, its ML estimator of its mean $\boldsymbol{\mu}^\prime$ and covariance $\boldsymbol{\sigma}^\prime$ is
\begin{align*}
\hat{\boldsymbol{\mu}}^\prime&=\frac{1}{N}\sum_{i=1}^{N}\mathbf{x}^\prime_i=\frac{1}{N}\{\sum_{\mathclap{i=1}}^{M}(\sqrt{\frac{N}{M}}\mathbf{x}_i + \hat{\boldsymbol{\mu}} - \sqrt{\frac{N}{M}}\hat{\boldsymbol{\mu}}) + \sum_{\mathclap{i=M+1}}^{N} \hat{\boldsymbol{\mu}}\}\\
&=\frac{1}{N}\{\sqrt{\frac{N}{M}}\sum_{\mathclap{i=1}}^{M}\mathbf{x}_i + M\hat{\boldsymbol{\mu}} - M\sqrt{\frac{N}{M}}\hat{\boldsymbol{\mu}} + (N-M)\hat{\boldsymbol{\mu}}\}\\
&=\frac{1}{N}\{\sqrt{\frac{N}{M}}M\hat{\boldsymbol{\mu}} + M\hat{\boldsymbol{\mu}} - M\sqrt{\frac{N}{M}}\hat{\boldsymbol{\mu}} + (N-M)\hat{\boldsymbol{\mu}}\}\\
&=\frac{1}{N}\{N\hat{\boldsymbol{\mu}}\} = \hat{\boldsymbol{\mu}},
\numberthis
\label{equ_outputmean}
\end{align*}
and
\begin{align*}
\hat{\boldsymbol{\sigma}}^\prime&=\frac{1}{N}\sum_{i=1}^{N}\{\mathbf{x}^\prime_i-\hat{\boldsymbol{\mu}}^\prime\}^2 = \frac{1}{N}\sum_{i=1}^{N}\{\mathbf{x}^\prime_i - \hat{\boldsymbol{\mu}}\}^2\\
&=\frac{1}{N}\sum_{i=1}^{M}\{\sqrt{\frac{N}{M}}\mathbf{x}_i + \hat{\boldsymbol{\mu}} - \sqrt{\frac{N}{M}}\hat{\boldsymbol{\mu}} - \hat{\boldsymbol{\mu}}\}^2 + \frac{1}{N}\sum_{\mathclap{i=(M+1)}}^{N}\{\hat{\boldsymbol{\mu}} - \hat{\boldsymbol{\mu}}\}^2\\
&=\frac{1}{N}\sum_{i=1}^{M}\{\sqrt{\frac{N}{M}}\mathbf{x}_i - \sqrt{\frac{N}{M}}\hat{\boldsymbol{\mu}}\}^2\\
&= \frac{1}{M}\sum_{i=1}^{M}\{\mathbf{x}_i - \hat{\boldsymbol{\mu}}\}^2 = \hat{\boldsymbol{\sigma}}.
\numberthis
\label{equ_outputvariance}
\end{align*}
\par Therefore, the proposed algorithm oversamples the original input dataset to a fixed number while keeping the ML estimation of mean and variance the same.


%

%
%
%
%
%

\ifCLASSOPTIONcaptionsoff
  \newpage
\fi




\bibliographystyle{IEEEtran}
\bibliography{IEEEabrv,mydatabase}
%

%





\end{document}